\documentclass[lettersize,journal]{IEEEtran}
\usepackage{amsmath,amsfonts}
\usepackage{algorithmic}
\usepackage{array}
\usepackage[caption=false,font=normalsize,labelfont=sf,textfont=sf]{subfig}
\usepackage{textcomp}
\usepackage{stfloats}
\usepackage{url}
\usepackage{verbatim}
\usepackage{graphicx}
\def\BibTeX{{\rm B\kern-.05em{\sc i\kern-.025em b}\kern-.08em
    T\kern-.1667em\lower.7ex\hbox{E}\kern-.125emX}}
\usepackage{balance}
\usepackage{multirow}

\begin{document}

\title{Privacy Challenges in Meta-Learning: An Investigation on Model-Agnostic Meta-Learning}

% \markboth{Journal of \LaTeX\ Class Files,~Vol.~18, No.~9, September~2020}%
% {How to Use the IEEEtran \LaTeX \ Templates}

\author{%
  \IEEEauthorblockN{%
    Mina Rafiei \IEEEauthorrefmark{1},
    Mohammadmahdi Maheri\IEEEauthorrefmark{2}
    and
    Hamid R. Rabiee\IEEEauthorrefmark{3} 
  }%
  
  \IEEEauthorblockA{\IEEEauthorrefmark{1} \textit{Sharif University of Technology}, Tehran, Iran, m.rafiei@sharif.edu  }%
  
  \IEEEauthorblockA{\IEEEauthorrefmark{2} \textit{Imperial College London}, London, United Kingdom, m.maheri23@imperial.ac.uk}%
  
  \IEEEauthorblockA{\IEEEauthorrefmark{3} \textit{Sharif University of Technology}, Tehran, Iran, rabiee@sharif.edu  }%
}

\maketitle

\begin{abstract}

Meta-learning involves multiple learners, each dedicated to specific tasks, collaborating in a data-constrained setting. In current meta-learning methods, task learners locally learn models from sensitive data, termed support sets. These task learners subsequently share model-related information, such as gradients or loss values, which is computed using another part of the data termed query set, with a meta-learner. The meta-learner employs this information to update its meta-knowledge. Despite the absence of explicit data sharing, privacy concerns persist. This paper examines potential data leakage in a prominent meta-learning algorithm, specifically Model-Agnostic Meta-Learning (MAML). In MAML, gradients are shared between the meta-learner and task-learners. The primary objective is to scrutinize the gradient and the information it encompasses about the task dataset. Subsequently, we endeavor to propose membership inference attacks targeting the task dataset containing support and query sets. Finally, we explore various noise injection methods designed to safeguard the privacy of task data and thwart potential attacks. Experimental results demonstrate the effectiveness of these attacks on MAML and the efficacy of proper noise injection methods in countering them.

\end{abstract}

\section{Introduction}
\label{sec:intro} 

Meta-learning, also known as "learning to learn," seeks to enhance the performance of machine learning (ML) methods by accumulating experience across multiple learning tasks~\cite {ThrunP98}. A meta-learner trains a model by iteratively consolidating knowledge from models trained on different tasks, encompassing various data types such as parameters, loss, and gradients. This process can be done in a centralized or distributed manner. In the distributed setting, each user wants to learn on their dataset while also wishing to benefit from the knowledge held in other datasets \cite{fallah2020personalized,lin2020collaborative}. Naturally, user privacy is essential in such settings.

Additionally, meta-learning is widely used in real-world distributed privacy-sensitive applications such as recommendation systems~\cite{lee2019melu,NIPS2017_51e6d6e6,metaselector,Task-adaptive}. Privacy concerns are particularly heightened when dealing with scenarios with limited training samples. This paper focuses on the Model-Agnostic Meta-Learning (MAML) algorithm, where the task learner shares gradients with the meta-learner. The gradient is computed from the loss of the network on a subset of data called the "query set." In contrast, the network parameters are obtained by adapting the initial parameters on another subset named the "support set." By sharing this gradient at each step of meta-learning, the meta-learner can access a gradient containing information about support and query sets. This leads to the central question of our work: \emph{Can meta-learner violate the privacy of task-learners?}

To address this question, we analyze the information contained in the shared gradients by task learners to identify potential privacy leakage. Subsequently, we propose two novel membership inference attacks (MIA) targeting the query and support sets. This attack is characterized by using a single gradient without altering the original process, making it a passive attack that goes unnoticed by the task learner. Our proposed malicious meta-learner highlights the risk of inferring members of task learners' training data involved in the meta-learning process, differing from malicious servers in federated learning~\cite{boenisch2021curious,fowl2022robbing}. The attack leverages a distinct gradient, computed by evaluating loss with task parameters on the query set but differentiating it by meta-learner parameters.

In summary, our contributions are as follows:
\begin{itemize}
    \item We analyze the shared gradient, illustrating the general challenges of attacking a task learner in the meta-learning process compared to federated learning. We delineate the distinctions between support and query sets concerning data leakage and privacy levels. This is the first in-depth investigation of data leakage from shared information in the MAML framework.

    \item  We introduce two new membership inference attacks tailored for MAML: one aimed at the support set and the other at the query set. These attacks leverage previous gradient-based reconstruction techniques but are customized to align with the distinct features of MAML. This sets them apart from earlier attacks tailored for federated learning, marking them as the pioneering privacy attacks developed for a meta-learning framework.
    
    \item  We investigate the efficacy of various noise injection techniques to safeguard privacy and their influence on the learning process. By introducing noise at different stages of the task learner process, we evaluate each method's effectiveness and identify the most suitable approach for various scenarios.

    \item Our experiments demonstrate that our designed attacks successfully infer the membership of individual data points in both the task query and support sets. Importantly, these attacks are passive, leaving the meta-learning process unaffected. Our paper also evaluates the privacy preservation methods against these attacks, exploring different noise injection strategies and their impact on privacy and the meta-learning process. The experiments demonstrate the ability to use different noise addition methods to prevent attacks without compromising the learning process.
    
\end{itemize}

In conclusion, our work provides a comprehensive understanding of the extent of privacy leakage in a MAML process, considering both support and query sets and the various parameters influencing the probability of successful attacks. Moreover, we offer insights into preventive measures to enhance confidence in data privacy. 

The paper is organized as follows: In Section \ref{sec:rw}, we delve into the background and review related work. Section \ref{sec:problem} elucidates the problem formulation and introduces relevant terminology. Our attack methodology and noise addition methods to counteract attacks are detailed in Section \ref{sec:methodology}. We unveil our experimental results in Section \ref{sec:expriment}. Finally, Section \ref{sec:conclusoin} summarizes our findings and provides concluding remarks.
\section{Related Work}
\label{sec:rw}

Privacy in machine learning and meta-learning is a critical and timely topic that we delve into in this paper. We start by discussing the attacks that aim to infer membership, a key privacy concern in machine learning. We then focus on understanding the privacy risks in the context of meta-learning.

\subsection{Privacy in Machine Learning}
\label{ml_privacy}

The privacy of the data and the models learned is essential in a machine learning (ML) process to learn a discriminative function based on an objective using a dataset. \cite{rigaki2020survey} have taxonomized privacy attacks based on different assets (such as the data or the model) under attack, actors in the ML process, and the knowledge and capabilities of the adversary. Different attacks, such as reconstruction attacks~\cite{gong2016you,he2019model,yang2019neural}, model extraction attacks~\cite{yang2019neural,milli2019model,orekondy2019knockoff,correia2018copycat,tramer2016stealing}, or inference attacks~\cite{AtenieseFMSVV13} are possible. One of the most popular attacks we focus on in this paper is the membership inference attack (MIA)~\cite{shokri2017membership}. In this attack, the adversary tries to guess whether a particular data point was used to train a model. MIAs can either be black-box or white-box~\cite{leino2020stolen}, rely on prediction outputs from models~\cite{shokri2017membership}, or logit values~\cite{song2021systematic}, or loss values~\cite{yeom2018privacy} and can be centralized~\cite{shokri2017membership, yeom2018privacy, long2018understanding} or decentralized~\cite{nasr2019comprehensive}. 

 To counter these potential attacks, a range of methods for privacy preservation have been proposed. Some approaches incorporate regularization, often in the form of dropout \cite{shokri2017membership}, \cite{SongSM19}, \cite{MelisSCS19}. In contrast, others aim to reduce the information shared with potential adversaries through techniques like gradient subset \cite{Shokriprivacy}, \cite{Yoon2020FederatedCL}, or gradient compression \cite{Konecn2016FederatedLS}. Another strategy involves differential privacy (DP), which formally defines privacy by ensuring that changing a single data point in the dataset should not significantly alter the algorithm's output \cite{Dwork2014}. DP has been successfully applied to deep learning, employing the concept of noise addition to gradients \cite{Shokriprivacy}.

Additionally, when the adversary is external to the learning process, alternative methods come into play. For instance, homomorphic encryption involves encrypting messages before transmission and decrypting them on the server \cite{Phong2018}, \cite{BosLN14}. Secure multiparty computation (SMC) is another approach, enabling participants in a learning process to compute functions with data shared among them in a way that prevents other parties from accessing the data \cite{Furukawa2017}, \cite{Mohassel2015}, \cite{araki2016high}.

\subsection{Privacy in Meta-Learning}
A more comprehensive introduction to meta-learning is provided in \S~\ref{subsec:meta}, which entails consolidating insights from experts across various tasks. Since each task is trained using distinct datasets, safeguarding the privacy of each task becomes paramount. Despite the significance of privacy in meta-learning, there has been limited research on this aspect. Initially, \cite{li2019differentially} pioneered the application of differential privacy \cite{abadi2016deep} in meta-learning. Subsequent efforts by \cite{hong2021learning,zhou2022task,dong2023padp} further explored meta-learning with a differential privacy guarantee. Notably, \cite{9538829} leveraged meta-learning to protect user-sensitive data by incorporating it into the support set. However, a thorough evaluation elucidating why the support set is deemed more private is lacking. Despite the growing interest in privacy within the context of meta-learning, there is currently no specific known attack against its privacy.

\section{Meta-Learning \& Avenues for Privacy Leakage}
\label{sec:problem}

In this section, we provide a meta-learning primer and describe avenues a malicious actor may exploit to learn private information.

\subsection{Meta-Learning Overview}
\label{subsec:meta}

Meta-learning aims to learn a model on various tasks with limited training data. It does so by learning and sharing knowledge from training individually on different tasks. In our setting, we define two actors: (a) task-learners who are responsible for learning from specific tasks and (b) a meta-learner who aggregates information from these task-learners.

\noindent{\bf Generic Formulation:} At each task-learner, the procedure is similar to conventional expectation risk minimization; a model (with specific parameters) will be learned by optimizing a suitable objective. Formally speaking, task-learner $i$ has a dataset $D_i = \{(x^i_j, y^i_j)_{j=1}^n\}$ and meta-knowledge $\omega$ (which we shall define next), and wishes to obtain parameters $\theta_i^{*}$ which minimizes a suitable loss function $\mathcal{L}_{task}$ i.e.,  
$\theta_i^{*} = \arg \min_{\theta_i} \mathcal{L}_{task}( \omega, \theta_i,D_i)$.
Candidates for the loss function include the negative log-likelihood and the cross-entropy loss~\cite {hospedales2021meta}. The meta-learner aims to learn knowledge associated with the task-learning process. This is collectively called {\em meta-knowledge} (referred to as $\omega$ earlier) and includes information about hyperparameters such as initialization weights, architecture, learning rate, etc.

Informally speaking, meta-learning is responsible for parameterizing the learning process to enhance the learning process for subsequent iterations. Formally, the meta-learner runs the following optimization (after $T$ tasks are learned by task-learners):
$\omega^{(*)} = \arg  \min_{\omega} \sum_{i=1}^{T} \mathcal{L}_{meta} ( \omega ,\theta_i ^{(*)} ,  D_i)$
where $\mathcal{L}_{meta}$ captures the meta-learner's objective. The meta-learning procedure operates iteratively, where task-learners learn their parameters locally before sharing them with the meta-learner for its optimization.

\textit{ In this work, we assume that the meta-knowledge is the parameters being optimized ( $\omega$ and $\theta$ are the same);  such an assumption is made in prior work by \cite{finn2017model} called  Model-Agnostic Meta-Learning (MAML) algorithm.}

\textbf{ Federated Meta Learning:} The aforementioned procedure can be done in a federated manner However, it is susceptible to privacy considerations; each task-learner shares (a) its locally computed parameters, as well as (b) their local dataset with the meta-learner. This may reveal sensitive information about the task-learner. In practice, the task learning process has two phases: (a) \textit{adaptation} and (b) \textit{validation}. In the adaptation phase, the task-learner uses meta-knowledge to learn a model on its (local and private) dataset. Then in the validation phase, some information from the learned model is computed (e.g., loss, gradient, etc.) to be used later in the meta-learning step.

To this end, each task-learner's dataset $D_i$ can be partitioned into disjoint sets $\{  D_i^s, D_i^q\}$. $D_i^s$ is referred to as the support set used for adaptation, while the query set $D_i^q$ is used in the validation phase. Thus, each task-learner computes the loss of its personal adapted model using $D_i^q$ and sends the gradient of the loss w.r.t the meta-knowledge to the meta-learner, where it updates its meta-knowledge (using this information). In this way, users do not share their personal dataset but share the common knowledge that helps improve learning. 

Formally speaking, at meta-learning step $t$, the task-learner runs the following optimization during the adaptation phase:

\begin{equation}
    \theta_{i}^{t+1} = \arg  \min_{\theta_i^t} \mathcal{L}_{task}( \omega^{t} , \theta_{i}^{t} ,D_i^s),
\end{equation}

and computes
\begin{equation}  
\nabla_{\omega^{t}}^i = \nabla_{\omega^{t}} \mathcal{L}_{task}(\omega^{t}, \theta_{i}^{t+1},  D_i^q)
\end{equation}
% \varun{check if this right; reference: MAML}
to share with the meta-learner during the validation phase. After learning $T$ tasks, the meta-learner updates the meta-knowledge as follows:
\begin{equation}
    \omega^{t+1} = \arg  \min_{\omega^{t}} \sum_{i=1}^{T} \mathcal{L}_{meta} (\nabla_{\omega^{t}}^i, \omega^{t})
\end{equation}

The learned $\omega_{t+1}$ is shared with the task-learners for subsequent iterations. 

\begin{figure*}[!t]
        \centering
        \includegraphics[scale=0.95]{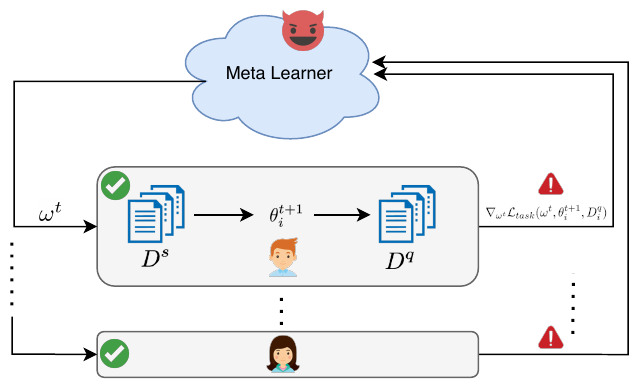}
        \caption{"The Federated Meta-Learning Process:  The tasks' data, highlighted in gray, are not directly shared with the meta-learner, but there may be data leakage in the shared gradient.}
        \label{fig:fed-meta}
\end{figure*}

This process is depicted in Figure \ref{fig:fed-meta}. Each task, at every step, receives meta-knowledge from the meta-learner. After the adaptation and validation steps, the task shares the gradient with the meta-learner. However, the query and support sets are not directly shared with the meta-learner and are supposed to remain private. Nevertheless, the shared gradient may contain information about them, potentially causing data leakage.

% \noindent{\bf Comparison to Federated Learning:} 

\subsection{Further Details about MAML Algorithm}
Previous studies have demonstrated that sharing gradients can lead to privacy breaches, resulting in both reconstruction attacks \cite{zhu2019deep} and membership inference attacks \cite{nasr2019comprehensive}. To examine privacy leakage in the MAML algorithm, a closer examination of the shared gradient between the task learner and meta learner is essential. This gradient differs from the conventional gradient in machine learning. According to \cite{nichol2018first}, the gradient can be computed as follows:

\begin{equation}
\begin{aligned}    
    g_{maml} & = \nabla_{\omega} \mathcal{L}_{task}(\omega, \theta,  D^q)  \\ & = \nabla_{\omega} \mathcal{L}( \omega,\mathcal{U}(\omega , D^s), D^q)  \\
     & = \nabla_{\omega} \mathcal{U}(\omega , D^s) \nabla_{\theta} \mathcal{L}( \omega, \theta  , D^q)
\end{aligned}
\end{equation}

Here, $\mathcal{U}$ represents the update function used in the adaptation phase, and $\mathcal{L}$ is the task loss function used instead of $\mathcal{L}_{task}$ for abbreviation. The update function $\mathcal{U}$ is given by:

\begin{equation}
    \mathcal{U}(\omega , D^s) = \omega + \nabla_{\omega_0} \mathcal{L}(\omega,\omega_0,D^s) + \nabla_{\omega_1} \mathcal{L}(\omega,\omega_1,D^s) + ...
\end{equation}

 where $\omega_0 = \omega$ and for $i > 0$, $\omega_i = \omega + \sum_{j=1}^{i} \nabla_{\omega_{j}} \mathcal{L}(\omega,\omega_j,D^s)$. To compute $g_{maml}$, we consider $\mathcal{U}$ up to the second phrase, essentially computing the second-order gradient for MAML. The resulting gradient is as follows:
 
 \begin{equation}
     \nabla_{\omega} \mathcal{L}_{task}(\omega, \theta,  D^q) = (I + \nabla_{\omega^2} \mathcal{L}(\omega,\omega,D^s) )\nabla_{\theta}  \mathcal{L}( \omega, \theta  , D^q)
 \end{equation}

This indicates that the shared gradient value between the task learner and meta learner includes the Hessian of the loss on the support set with respect to $\omega$ and the gradient of the loss on the query set with respect to $\theta$, potentially leading to data leakage from both the support and query sets.

The authors of \cite{finn2017model} also introduced another variant of MAML called first-order MAML. In this version, the gradient is computed using a first-order approximation to reduce computation costs. The gradient, in this case, is given by:

\begin{equation}
     \nabla_{\omega} \mathcal{L}_{task}(\omega, \theta,  D^q) =  \nabla_{\theta} \mathcal{L}( \omega, \theta  , D^q)
 \end{equation}

 As evident, in the first-order approximation, there is no longer any term directly involving support data. Consequently, it is anticipated that it will be harder to access the support data in this variant of the MAML algorithm.

\subsection{Privacy Leakage in MAML}
\label{sec:Attacks on MAML}
As demonstrated earlier, each task's gradient sent to the meta-learner incorporates information about both the support and query sets. Our objective is to utilize this gradient to devise an attack. However, before proceeding, a fundamental question arises: is it feasible to retrieve both the support and query sets solely based on the gradient? This question will be explored further in this section. To undertake this investigation, we will delve into the next proposition:

\newtheorem{theorem}{Theorem}[section]
\newtheorem{proposition}[theorem]{Proposition}
\begin{proposition}
\label{prop:prop1}
Consider a neural network architecture featuring a biased fully-connected layer, preceded by a (possibly unbiased) fully-connected layer. This network undergoes training using the MAML algorithm. Let's focus on a particular task, denoted as $i$, associated with data $D = \{ D^s, D^q \}$, consisting of a support set ($D^s$) and a query set ($D^q$). In iteration $t$, with the meta-learner's meta-knowledge represented as $\omega$, task $i$ computes the gradient as $g = \nabla_{\omega} \mathcal{L}(\omega, \theta, D^q)$ with the intention of sharing this gradient with the meta-learner.
If there exists dataset $M^s \neq D^s$ for which $I + \nabla_{\omega^2} \mathcal{L}(\omega,\omega,M^s)$ is invertible and $ g (I + \nabla_{\omega^2} \mathcal{L}(\omega,\omega,M^s) )^{-1} $ contains at least one non-zero entry in each row, then there exists $M^q \neq D^q$ such that replacing $D$ with $M = \{ M^s , M^q \}$ generates the same gradient $g$.
\end{proposition}

\begin{IEEEproof}
When $M^s$ is used in the task adaptation phase, it leads to an adapted parameter $\theta^\prime$ instead of $\theta$. To maintain the same gradient $g$, we need to choose $M^q$ such that: :
 \begin{equation}
     \begin{aligned}
         g &= \nabla_{\omega} \mathcal{L}(\omega, \theta, D^q)
         \\
         &= \nabla_{\omega} \mathcal{L}(\omega, \theta^{\prime}, M^q)
         \\
         &= (I + \nabla_{\omega^2} \mathcal{L}(\omega,\omega,M^s) )\nabla_{\theta^{\prime}}  \mathcal{L}( \omega, \theta^{\prime}  , M^q)
     \end{aligned}
 \end{equation}

The invertibility of $I + \nabla_{\omega^2} \mathcal{L}(\omega,\omega,M^s)$ allows us to express $\nabla_{\theta^{\prime}} \mathcal{L}( \omega, \theta^{\prime} , M^q)$ as $g^\prime = g (I + \nabla_{\omega^2} \mathcal{L}(\omega,\omega,M^s) )^{-1}$.

Now by Utilizing the proposition from \cite{geiping2020inverting} (Data Reconstruction from Gradients), we can reconstruct $M^q$ from gradient $g^\prime$ by ensuring that $g^\prime$ contains at least one non-zero entry at each layer. to be more clear we have $\nabla_{\theta^{\prime}} \mathcal{L}( \omega, \theta^{\prime} , M^q) = g^\prime$ and we have $\theta ^\prime$ so we can reconstruct the input and get $M^q$.
    
\end{IEEEproof}

However, a more nuanced inquiry arises: do these two conditions mentioned in proposition \ref{prop:prop1} hold for neural networks? Providing a general answer to this question is challenging, as it heavily relies on the specific architecture of the network. Yet, we address this uncertainty in the appendix \ref{app1} by demonstrating the existence of networks for which both conditions are satisfied. Consequently, we assert that, in general, networks exist wherein the shared gradient in MAML lacks sufficient information about the query and support sets. Consequently, extracting them solely from the gradient is not always feasible, necessitating additional information.

Up to this point, we have established that attacking the task dataset using the shared gradient in the MAML algorithm presents greater challenges compared to what has been done before in the simple learning setting \cite{zhu2019deep}. \textit{However, is the difficulty of attacking the support set different from that of the query set?} Given that the gradient is computed using the loss on the query set, it suggests that the potential leakage of support set data could be less. This concept is utilized in the privacy preservation method proposed by the authors of \cite{9538829} for federated learning with meta-learning algorithms, where sensitive data is strategically placed in the support set to enhance protection against attacks.

Nevertheless, we aim to delve deeper into this matter. As previously demonstrated, the shared gradient comprises two multiplied terms: one term containing the gradient on the query set and the other term containing the hessian on the support set. Given that the gradient term inherently holds more information about the data compared to the hessian term, we can infer that the shared gradient contains more information about the query set in comparison to the support set.

Moreover, prior research indicates that having the gradient of loss and network weights allows for the reconstruction of the input to the network. However, there is no existing method to reconstruct the support set solely from having the hessian, thereby heightening the complexity of the attack on the support set.

\section{Attack Methodology}
\label{sec:methodology}

\subsection{Threat Model}
\label{subsec:threat}

The objective of a privacy adversary in the meta-learning ecosystem may vary. For example, the adversary may wish to {\em infer membership} of a particular data point in a specific task-learner's dataset, or it may aim to {\em reconstruct} the data used in the information shared from the task-learner.
Such an adversary may be {\em passive} and follow protocol specifications or be {\em active} and subvert them when necessary. The attack itself may be {\em local} (when targeting a particular task-learner) or {\em global} (when targeting all task-learners). 

This work considers a scenario where the meta-learner is the privacy adversary. Thus, it can access all information shared by each task-learner and the meta-knowledge. This entity is also assumed to be passive, and this attack will not affect the meta-learning process. The adversary aims to perform membership inference locally, i.e., to determine the presence or absence of a data point in the training data of one task learner.

As explained earlier, the adversary's objective is to infer the membership of a data point the task learner uses. We formalize this requirement through an MI game (between a challenger and the adversary), as commonly done~\cite {yeom2018privacy,jagielski2022combine}. Borrowing insight from~\cite{jagielski2022combine}, we define the game from the challenger's perspective as follows:

\begin{enumerate}
\item Sample $a \in \{1,\cdots,T\}$ and $b \in \{0,1\}$ uniformly at random. $b$ captures whether the sample is in the training set, and $a$ captures which task learner it comes from.
\item If $b=0$, sample $(x,y) \sim \mathcal{D}$ (the distribution of data). Else, sample $(x,y) \sim D_a$ (the task data).
\item Run the meta-learning process as described earlier, and share the corresponding information (e.g., $\{ g_1, \cdots,g_T \}$ and associated meta-knowledge ($\omega$) with the adversary. 
\end{enumerate}

In such a game, the adversary (i.e., malicious meta-learner) wins if he can correctly guess both $b$ and $a$ in Golabl attacks. For locally targeted attacks, guessing $b$ for a specified $a$  suffices.

\subsection{Intuition}
\label{subsec:intuition}
In planning a passive attack, we rely solely on the information shared with the meta-learner, which is encapsulated in the gradient. As previously mentioned, the gradient encompasses support and query sets information. However, recognizing that the type and volume of information within the gradient from each set vary, we can formulate distinct attacks tailored for each query and support set. This strategy acknowledges the unique characteristics of the information embedded in the gradient from these different sets, enabling the design of targeted attacks based on the nature of the data.

Our approach involves estimating the gradient of the query set through the task gradient to target the query set. This estimated query set gradient can then be employed for the attack. However, attacking the support set proves to be more challenging since the direct gradient of the support set is not explicitly embedded in the gradient value. Further exploration of this complexity will be undertaken in the subsequent sections, where we will delve into more intricate details.

\subsection{Membership Inference Attack on Query Data }
In previous research \cite{geiping2020inverting}, it has been demonstrated that given the gradient of a neural network loss on the input with respect to its weights, it is possible to reconstruct the input. To reconstruct the query set $D_i^q$ belonging to task learner $i$ at iteration $t$ from its gradient, we ideally need $\nabla_{\theta_i^t} \mathcal{L}(\omega^t, \theta_i^t, D_i^q)$ and $\theta_i^t$—where $\theta_i^t$ represents the adapted parameters of task learner $i$ at iteration $t$, and $\omega^t$ is the meta-knowledge at that iteration.

However, what we have is $\nabla_{\omega^t} \mathcal{L}(\omega^t, \theta_i^t, D_i^q)$ with $\omega^t$. To proceed with an attack, if we can estimate $\theta_i^t$ using $\omega^t$, it becomes feasible to leverage the shared gradient. 
% The rationale is that if $\omega^t$ and $\theta_i^t$ are close, it implies that $\mathcal{L}(\omega,\omega,D^s_i)$ was near zero. Consequently, $\nabla_{\omega^2}\mathcal{L}(\omega,\omega,D^s_i)$ is negligible, and $\nabla_{\theta_i^t} \mathcal{L}(\omega^t, \theta_i^t, D^q)$ can be approximated by $\nabla_{\omega^t} \mathcal{L}(\omega^t, \theta_i^t, D^q)$, as indicated by equation \ref{}.
But is this assumption rational? The purpose of meta-learning, particularly through algorithms like MAML, is to facilitate task-learners' adaptation with minimal steps. The meta-knowledge ($\omega^t$) is designed to encode information that enables quicker adaptation of task-specific parameters ($\theta_i^t$) during the learning process.
Given this design philosophy, it's reasonable to expect that the difference between $\omega^t$ and $\theta_i^t$ is not substantial. The meta-knowledge is crafted to capture shared patterns or knowledge across tasks, allowing task learners to adapt efficiently. Consequently, if $\omega^t$ and $\theta_i^t$ are close, the meta-knowledge has facilitated a relatively swift adaptation for the specific task at iteration $t$. This expectation forms a rational basis for the assumption made in the context of the attack strategy.

The attack strategy relies on a reconstruction attack, and a distinctive approach is employed to conduct a membership inference attack based on this reconstruction. Instead of solely comparing the reconstructed data with the target data, an additional step is introduced to enhance the efficiency of the reconstruction attack.

In this method, the target data serves as the prior input for the reconstruction attack. Following the reconstruction process, a comparison is made between the initial input and the last reconstructed image. If the observed changes between the initial input and the final reconstructed image are minimal, it suggests that the target data may indeed be a member of the query data. Conversely, This approach adds a layer of scrutiny by considering the extent of changes during the reconstruction, providing a nuanced assessment of the membership inference.

Let's consider the objective function used in \cite{geiping2020inverting} to delve further into the details. Assuming the target attack data is denoted as $x_0$, and we aim to assess its membership in task $i$ query set $D_i^q$ using the shared gradient $g = \nabla_{\omega^t} \mathcal{L}(\omega^t, \theta_i^t, D_i^q)$ obtained in iteration $t$, the optimization problem is formulated as follows:

\begin{equation}
    x^* = \arg_x \min 1-l(\nabla_{\omega} \mathcal{L}(\omega^t,\omega^t,x),g) + \alpha TV(x)
\end{equation}

\begin{figure*}[!t]
        \centering
        \includegraphics[scale=0.5]{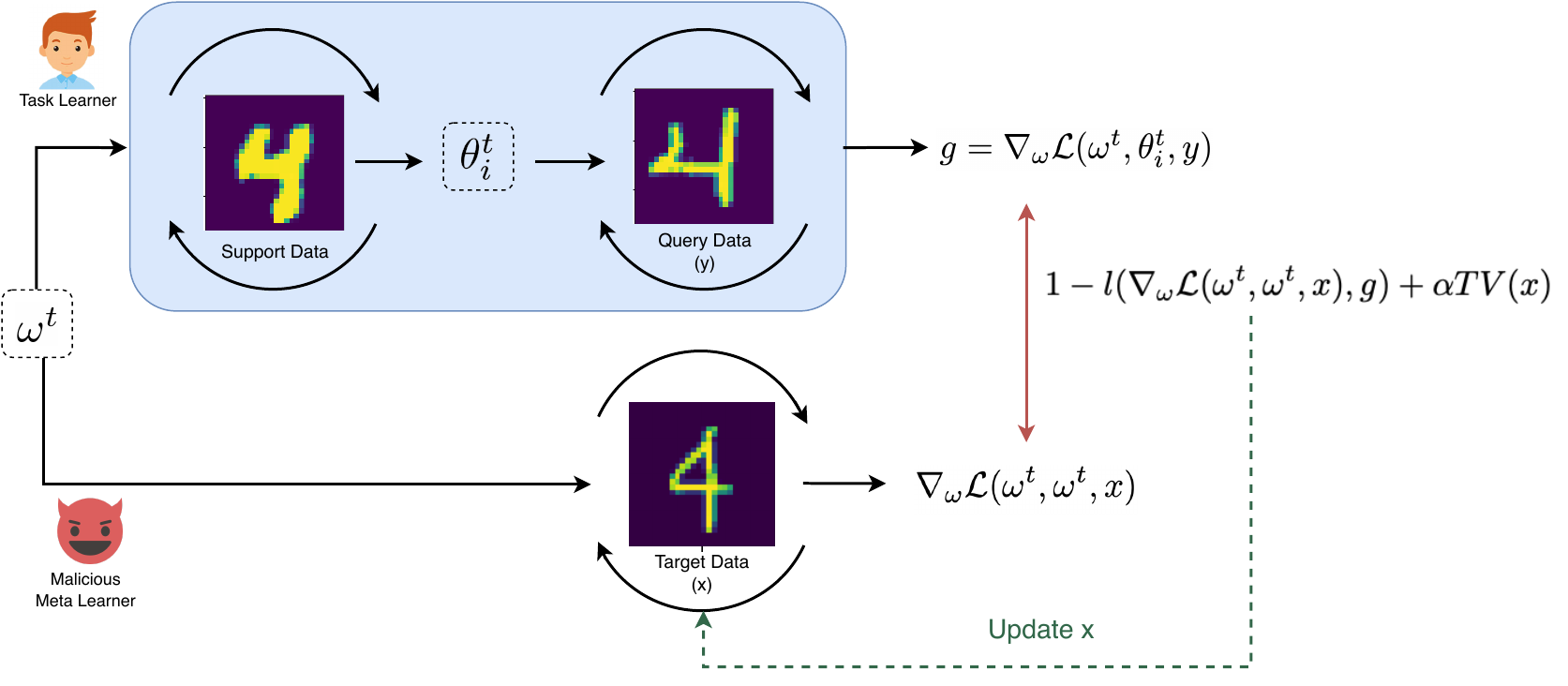}
        \caption{The Query Reconstruction Process Utilizing the Target Image as Prior Subsequently, the reconstructed query and its prior are compared to infer the membership of the target image in the task query set.}
        \label{fig:query-attack}
\end{figure*}

Here, $l$ represents the cosine similarity, and $\text{TV}$ stands for total variation, as introduced in \cite{rudin1992nonlinear} and utilized in \cite{geiping2020inverting} as a prior for image reconstruction. The process is depicted in Figure \ref{fig:query-attack} where the target image $x_0$ is considered as the prior for data reconstruction from the gradient.

After the reconstruction, an assessment of membership is conducted by comparing the original data $x_0$ with the reconstructed data $x^*$. For this comparison, we utilize the Structural Similarity Index (SSIM) metric \cite{wang2004image}. SSIM is a perception-based metric that takes into account structural information in images by considering inter-dependencies among pixels. This is in contrast to metrics such as Mean Squared Error (MSE) or Peak Signal-to-Noise Ratio (PSNR), which rely solely on absolute errors.  
So if the $SSIM(x_0,  x^*)$ is above  a predefined threshold $d$, the inference is made that $x_0$ was a member of $D^q$; otherwise, it is inferred that $x_0$ was not a member. The threshold $d$ is determined based on the data distribution, utilizing a small subset of the data as a validation set.

It's worth noting that the above equations were initially formulated under the assumption that $D^q$ consists of a single data point. However, the extension to accommodate a batch of data in $D^q$ is straightforward. In such cases, $x_0$ is treated as one of the image priors in the data batch, while the remaining data points require no prior and are represented by noisy images.

\subsection{Membership Inference Attack on Support Data }

To adapt the attack methodology for the support set, it's crucial to acknowledge the limitations posed by having only the Hessian of the loss on the support data, which may not provide sufficient information about the gradient. Consequently, a modification in the approach is necessary.

As discussed earlier, attacking the support data without access to the query data can be challenging in certain scenarios. Moreover, in the interest of designing robust defenses against potential privacy leakages, it is valuable to develop attacks with stronger adversaries. The rationale is to create adversarial scenarios that are more challenging and then fortify the system to withstand such attacks. This approach is akin to the strategy employed in \cite{balle2022reconstructing}.

In this context, we assume that the adversary aiming to attack task $i$'s support set $D^s_i$ possesses access to the corresponding task query data $D^q_i$. This assumption broadens the scope of the adversary's knowledge, and the attack methodology can be adapted to leverage information from the query set for a more potent adversarial strategy.

With access to the query set, a similar attack strategy can be designed for attacking the support set. The process is akin to the previous attack, but now the adversary doesn't have direct access to the gradient of the loss on the support set. Instead, the adversary possesses the gradient shared by the task learner, which is computed based on the loss on the query set.

The attack unfolds as follows: assuming the attacker aims to infer the membership of data point $x_0$ in the support set $D_i^s$ from the gradient $g = \nabla_{\omega} \mathcal{L}(\omega^t, \theta^t_i, y)$ shared by task $i$ at iteration $t$ computed on task query data $y$. The attacker emulates a process similar to that of the task learner, with the objective of obtaining the gradient by assuming that $x$ is part of the support data for task $i$. In doing so, the attacker seeks to optimize $x$ using a similar cost function as in the reconstruction of $D_i^s$. The adaptation phase involves obtaining $\theta_i^{\prime t}$ by utilizing $x$, and subsequently computing $\nabla_{\omega} \mathcal{L}(\omega^t, \theta_i^{\prime t}, y)$. The optimization is then performed as follows:

\begin{equation}
     x^* = \arg_x \min 1-l(\nabla_{\omega} \mathcal{L}(\omega^t,\theta_i^{\prime t},y),g) + \alpha TV(x) 
\end{equation}

\begin{figure*}[!t]
        \centering
        \includegraphics[scale=0.5]{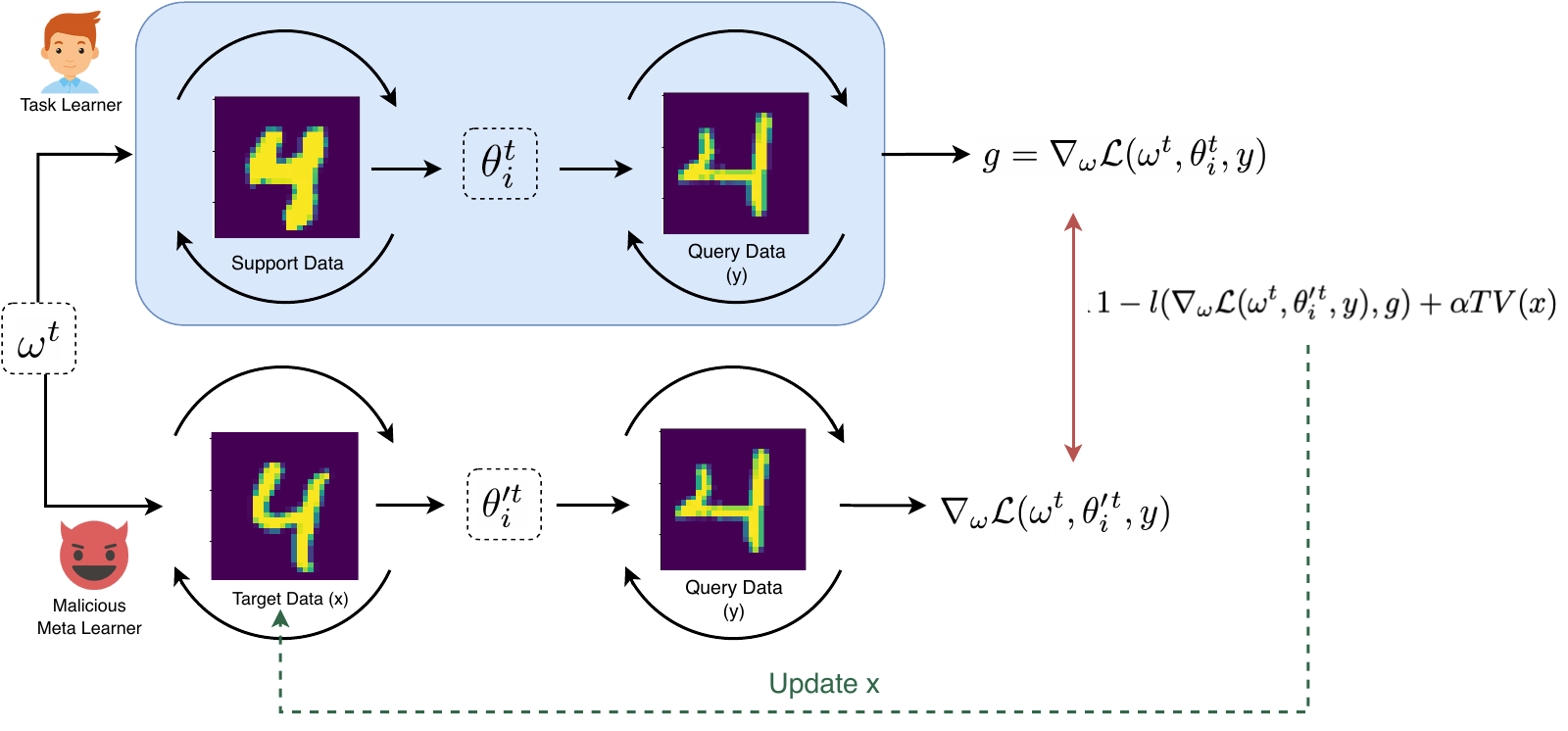}
        \caption{The Support Reconstruction Process Utilizing the Target Image as Prior:  Subsequently, the reconstructed support and its prior are compared to infer the membership of the target image in the task support set.}
        \label{fig:support-attack}
\end{figure*}

The process is depicted in Figure \ref{fig:support-attack} where the support data is reconstructed using the target image as the prior. As depicted in the figure, the attacker mimics the actions of the task learner to generate a gradient. Subsequently, by comparing this gradient to the ground truth gradient, the optimization process aims to recreate the original input.

After computing $x^*$, the inference of membership for data point $x_0$ involves calculating the $SSIM(x_0, x^*)$ and comparing it to a predefined threshold $d$. If the similarity is above  $d$, it can be concluded that $x_0$ is a member; otherwise, it is inferred that $x_0$ is not a member. Similar to the query attack, the optimization process can be extended to a batch of data, assuming that $x_0$ is one of the data points in the batch. This allows for a more comprehensive assessment of membership inference in a broader context.

\subsection{How to protect MAML from Attacks?}
The paramount objective in studying privacy leakages across various learning processes and designing counterattacks is to develop robust approaches to safeguard against different adversaries. As research on privacy attacks has grown, corresponding efforts have been directed towards privacy-preserving mechanisms. \cite{mireshghallah2020privacy} categorizes these mechanisms based on the phase during which the attack occurs—specifically, into data aggregation, training, and inference phases. Each phase poses distinct challenges for privacy preservation, necessitating tailored protection strategies.

In the context of meta-learning, our primary focus is on the training phase. The adversary in this scenario is the meta-learner itself, distinguishing it from data aggregation or inference phases. For the training phase, privacy-preserving mechanisms are classified into three categories: differential privacy, Homomorphic Encryption, and Secure Multi-Party Computation (SMC). Given that the meta learner as one of the parties serves as the adversary, the use of Encryption or SMC methods to protect against attacks is not feasible.

Therefore, the most effective strategy for ensuring the meta-learner's safety from privacy attacks in the training phase is to employ differential privacy methods. Differential privacy involves introducing noise into the learning process to protect sensitive information. \cite{li2019differentially} has proposed a differential private meta-learning process that incorporates the addition of noise to the gradient during the adaptation phase. However, this approach has yet to be rigorously tested against meta-learning attacks to evaluate its utility and effectiveness. Further research and experimentation are needed to assess the practicality and robustness of this differential privacy mechanism in the context of meta-learning.

In the context of studying the utility of adding noise to different parts of a meta-learning process, the aim is to protect the task learner from potential attacks on the support and query sets that the meta-learner could orchestrate. Hence, we can introduce noise at various stages to determine the most effective method for each attack. The noise, denoted as $\epsilon$, can be introduced into the gradient computation process at four distinct locations: Support Data, Adaptation Gradient, Query Data, and Validation Gradient.
\begin{itemize}
\item \textbf{Adding Noise to Support Data:}

Approach: Directly add noise to the support data before starting the adaptation phase in the task-learner.

Implementation: $\text{Noisy Support} = \text{Original Support} + \epsilon$, where $\epsilon \sim \text{Gaussian}(\mu, \sigma)$ and $\mu$ is the mean, $\sigma$ is the variance of the Gaussian noise.

\item \textbf{Adding Noise to Adaptation Gradient:}

Approach: Intervene in the gradient computation process in the adaptation phase by adding noise to the gradient of the computed loss.

Implementation: $\text{Noisy Gradient} = \text{Original Gradient} + \epsilon$, where $\epsilon \sim \text{Gaussian}(\mu, \sigma)$.

\item \textbf{Adding Noise to Query Data:}

Approach: Directly add noise to the query data before starting the validation phase in the task-learner.

Implementation: $\text{Noisy Query} = \text{Original Query} + \epsilon$, where $\epsilon \sim \text{Gaussian}(\mu, \sigma)$ and $\mu$ is the mean, $\sigma$ is the variance of the Gaussian noise.

\item \textbf{Adding Noise to Validation Gradient:}

Approach: Intervene in the gradient computation process in the validation phase by adding noise to the gradient of the computed loss.

Implementation: $\text{Noisy Gradient} = \text{Original Gradient} + \epsilon$, where $\epsilon \sim \text{Gaussian}(\mu, \sigma)$.
    
\end{itemize}

In a meta-learning process, the task learner typically has two gradient computation steps: one during the adaptation phase on support data and the other during the validation phase by query data. Therefore, the noise addition can be implemented at either or both of these phases to enhance privacy protection and mitigate the risk of privacy leakages. This strategic use of noise can fortify the meta-learning process against potential attacks from the meta-learner on the support and query sets.

\section{Experiments}
\label{sec:expriment}

Our well-structured experimental approach aligns with the typical methodology for validating attacks and evaluating privacy preservation mechanisms. Let's break down the two main sections and our goals in each section:

\begin{enumerate}
    \item \textbf{Validation of the Attack}:
    \begin{itemize}
        \item \textbf{Effectiveness}: In the first step, we need to evaluate how well the attack succeeds in breaching the privacy of the meta-learning process.
        
        \item \textbf{Sensitivity}: We analyze the sensitivity of the attack to variations in meta-learning algorithm configurations and task data. Understanding the robustness and adaptability of the attack is crucial.

        \item \textbf{Efficiency}: We also check the attack's efficiency by considering the time required and scalability factors.
    \end{itemize}

    \item \textbf{Validation of the Privacy Preservation Mechanism}:
    \begin{itemize}
        \item \textbf{Success in Preventing Attack}: We evaluate how well the privacy preservation mechanism, based on adding noise to different parts of the meta-learning process, mitigates or prevents the attack.
        \item \textbf{Impact on Meta Learning}: We ensure that adding noise does not significantly damage the main meta-learning process. We evaluate noise ranges that can prevent attacks and not damage learning processes.
    \end{itemize}
\end{enumerate}

 We comprehensively assess the attack and the proposed privacy preservation mechanism by systematically conducting experiments in these two sections. This approach will contribute valuable insights into the effectiveness and practicality of our proposed solutions.

\subsection{Experimental Setup}

We assess our attack utilizing two publicly available datasets: MNIST~\cite{deng2012mnist} and Fashion-MNIST~\cite{xiao2017fashion}. The code is implemented in Python using the PyTorch library and will be made publicly available after the paper is published. Our experimentation is conducted in the context of few-shot learning using the MAML algorithm. In this setting, the objective is to acquire classification capabilities with minimal data samples. For each dataset, we randomly select $2 \times s$ images from a total of $w$ classes, where $w$ denotes the overall number of classes, and $s$ represents the number of shots. Each task-learner is provided with support and query sets, both containing $s$ samples from $w$ classes, and these sets are ensured to be disjoint. Commonly used meta-learning architectures serve as the classifiers in our experiments. To find the best hyperparameters for the attack, we separate a portion of the data as the validation set and then tune the hyperparameters accordingly.

\subsection{Validation of Proposed Attack}

\subsubsection{Effectiveness}
\begin{figure}[!t]
        \centering
        \includegraphics[scale=0.35]{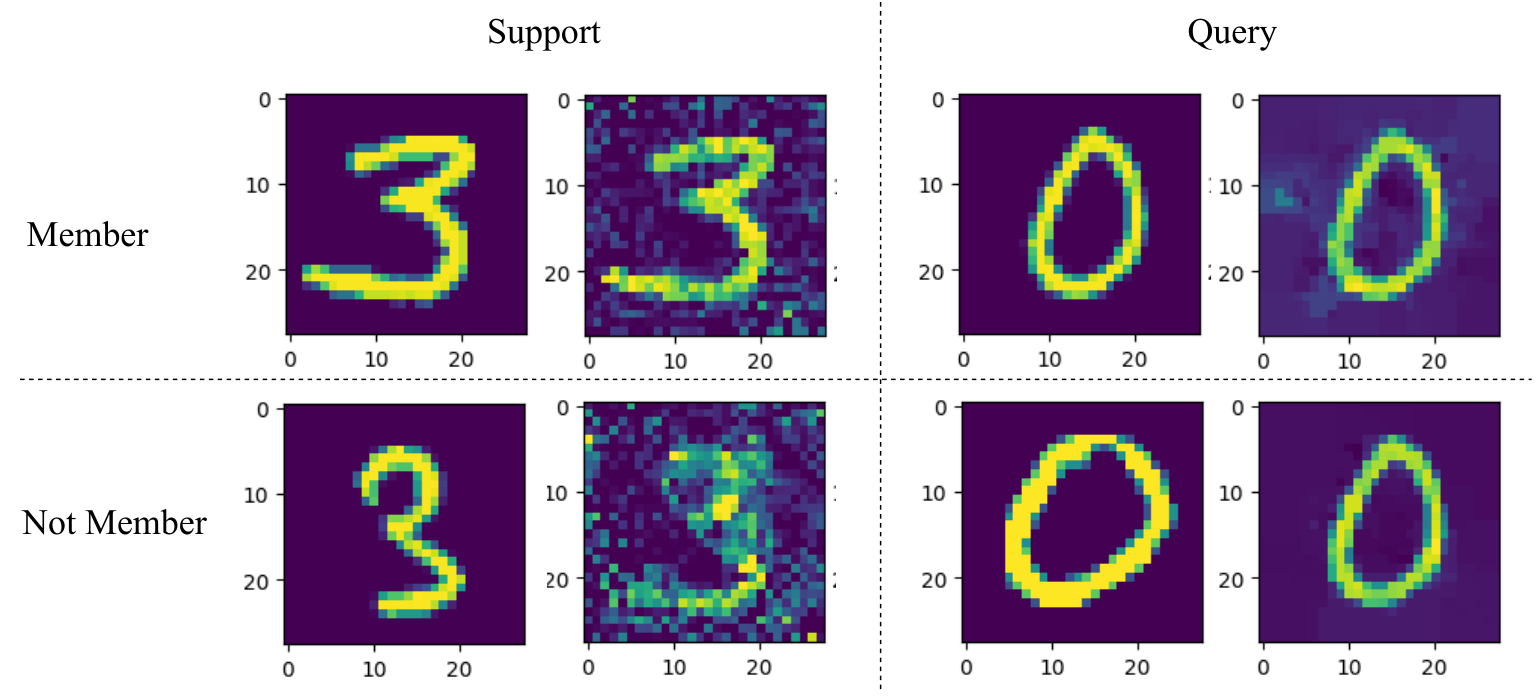}
        \caption{Attack sample runs. The right image in each part represents the output, and the left image is the attack input. As observed, when the input is a member, the output is similar to it. However, when it is not a member, the attack iterations cause changes, making it similar to the shape of the original member image. Thus, we can differentiate between these two by comparing the input and output classes.}
        \label{fig:sample-attack}
\end{figure}

In Figure \ref{fig:sample-attack} we present a sample illustrating how the attack operates. The left image in each part of the figure represents the input to the attack, while the right image shows the attack output. When the input image is a member of the task dataset, we observe that the output is similar to it. However, when the input is not a member of the task dataset, we expect that after attack iterations, the output changes and becomes similar to a real task member. This allows us to differentiate between member and non-member classes by comparing the input and output of the attack. The difference in output image quality between the query and support attack also supports the idea that attacking the support set is harder than the query set due to less information embedded in the gradient about it compared to the query set.

To evaluate the effectiveness of the attack, we employ standard classification metrics, including accuracy, recall, and precision. The goal of the attack is to predict whether the selected data belongs to the specific task or not. We conduct the experiment $n$ times, with half of the trials involving target data selected from a particular task and the other half randomly chosen from the data distribution, representing non-membership data. The results for the support attack are presented in Table \ref{tab:sup-attack}, while Table \ref{tab:query-attack} displays the outcomes for the query attack. The results demonstrate the success of the attack in distinguishing between member and non-member data.  

\begin{table*}[!t]

\begin{center}
\caption{Support Attack Results: The findings indicate that the attack is capable of inferring the membership of data within the task support dataset.}
\begin{tabular}{|c|c|ccc|ccc|}
\hline
\multirow{2}*{Ways} & \multirow{2}*{Shots} & \multicolumn{3}{c|}{MNIST}                                              & \multicolumn{3}{c|}{Fashion-MNIST}                                      \\ \cline{3-8} 
                      &                        & \multicolumn{1}{c|}{Accuracy} & \multicolumn{1}{c|}{Recall} & Precision & \multicolumn{1}{c|}{Accuracy} & \multicolumn{1}{c|}{Recall} & Precision \\ \hline
\multirow{3}{*}{3}    & 1                       & \multicolumn{1}{c|}{0.79}     & \multicolumn{1}{c|}{0.72}   & 0.84      & \multicolumn{1}{c|}{0.79}     & \multicolumn{1}{c|}{0.62}   & 0.94      \\ \cline{2-8} 
                      & 2                      & \multicolumn{1}{c|}{0.68}     & \multicolumn{1}{c|}{0.74}   & 0.66      & \multicolumn{1}{c|}{0.7}     & \multicolumn{1}{c|}{0.52}   & 0.81      \\ \cline{2-8} 
                      & 3                      & \multicolumn{1}{c|}{0.61}     & \multicolumn{1}{c|}{0.54}   & 0.63      & \multicolumn{1}{c|}{0.59}     & \multicolumn{1}{c|}{0.26}   & 0.76      \\ \hline
\multirow{3}{*}{5}       & 1                    & \multicolumn{1}{c|}{0.91}     & \multicolumn{1}{c|}{0.98}   & 0.86      & \multicolumn{1}{c|}{0.7}      & \multicolumn{1}{c|}{0.48}   & 0.86      \\ \cline{2-8} 
                      & 2                      & \multicolumn{1}{c|}{0.69}     & \multicolumn{1}{c|}{0.96}   & 0.62      & \multicolumn{1}{c|}{0.61}     & \multicolumn{1}{c|}{0.26}    & 0.87      \\ \cline{2-8} 
                      & 3                      & \multicolumn{1}{c|}{0.61}     & \multicolumn{1}{c|}{0.86}   & 0.57      & \multicolumn{1}{c|}{0.58}     & \multicolumn{1}{c|}{0.16}   & 1         \\ \hline
\multirow{3}{*}{7}       & 1                    & \multicolumn{1}{c|}{0.71}     & \multicolumn{1}{c|}{0.96}   & 0.64      & \multicolumn{1}{c|}{0.73}     & \multicolumn{1}{c|}{0.5}    & 0.93      \\ \cline{2-8} 
                      & 2                      & \multicolumn{1}{c|}{0.56}     & \multicolumn{1}{c|}{0.66}   & 0.55      & \multicolumn{1}{c|}{0.59}     & \multicolumn{1}{c|}{0.2}    & 0.91      \\ \cline{2-8} 
                      & 3                      & \multicolumn{1}{c|}{0.51}     & \multicolumn{1}{c|}{0.66}   & 0.51      & \multicolumn{1}{c|}{0.58}     & \multicolumn{1}{c|}{0.24}   & 0.75        \\ \hline
\end{tabular}

 \label{tab:sup-attack}
\end{center}
\end{table*}

\begin{table*}[!t]
\begin{center}
\caption{Query Attack Results: The findings indicate that the attack is capable of inferring the membership of data within the task query dataset.}
\begin{tabular}{|c|c|ccc|ccc|}
\hline
\multirow{2}*{Ways} & \multirow{2}*{Shots} & \multicolumn{3}{c|}{MNIST}                                              & \multicolumn{3}{c|}{Fashion-MNIST}                                      \\ \cline{3-8} 
                      &                        & \multicolumn{1}{c|}{Accuracy} & \multicolumn{1}{c|}{Recall} & Precision & \multicolumn{1}{c|}{Accuracy} & \multicolumn{1}{c|}{Recall} & Precision \\ \hline
\multirow{3}{*}{3}    & 1                      & \multicolumn{1}{c|}{0.92}     & \multicolumn{1}{c|}{0.98}   & 0.88      & \multicolumn{1}{c|}{0.78}     & \multicolumn{1}{c|}{0.96}   & 0.71     \\ \cline{2-8} 
                      & 2                      & \multicolumn{1}{c|}{0.71}     & \multicolumn{1}{c|}{0.52}   & 0.84      & \multicolumn{1}{c|}{0.73}     & \multicolumn{1}{c|}{0.66}   & 0.77      \\ \cline{2-8} 
                      & 3                      & \multicolumn{1}{c|}{0.59}     & \multicolumn{1}{c|}{0.24}   & 0.80      & \multicolumn{1}{c|}{0.62}     & \multicolumn{1}{c|}{0.36}   & 0.75      \\ \hline
\multirow{3}{*}{5}    & 1                      & \multicolumn{1}{c|}{0.95}     & \multicolumn{1}{c|}{0.98}   & 0.92      & \multicolumn{1}{c|}{0.87}     & \multicolumn{1}{c|}{1}    & 0.79      \\ \cline{2-8} 
                      & 2                      & \multicolumn{1}{c|}{0.69}     & \multicolumn{1}{c|}{0.46}   & 0.85      & \multicolumn{1}{c|}{0.7}      & \multicolumn{1}{c|}{0.54}   & 0.79      \\ \cline{2-8} 
                      & 3                      & \multicolumn{1}{c|}{0.69}     & \multicolumn{1}{c|}{0.38}   & 1         & \multicolumn{1}{c|}{0.69}     & \multicolumn{1}{c|}{0.44}   & 0.87      \\ \hline
\multirow{3}*{7}    & 1                      & \multicolumn{1}{c|}{0.93}     & \multicolumn{1}{c|}{1}   & 0.88      & \multicolumn{1}{c|}{0.82}     & \multicolumn{1}{c|}{0.96}   & 0.75      \\ \cline{2-8} 
                      & 2                      & \multicolumn{1}{c|}{0.74}     & \multicolumn{1}{c|}{0.52}   & 0.93      & \multicolumn{1}{c|}{0.67}     & \multicolumn{1}{c|}{0.44}   & 0.81      \\ \cline{2-8} 
                      & 3                      & \multicolumn{1}{c|}{0.67}     & \multicolumn{1}{c|}{0.36}   & 0.95      & \multicolumn{1}{c|}{0.7}      & \multicolumn{1}{c|}{0.46}   & 0.88      \\ \hline
\end{tabular}
\end{center}

\label{tab:query-attack}
\end{table*}

\subsubsection{Sensitivity}
We analyze the sensitivity of the attack to different meta-learning parameters. Examining the impact of each parameter on the attack provides insights into what measures task learners can take to decrease the probability of being successfully attacked. One crucial parameter is the task data size. The initial intuition is that increasing the data size will decrease the attack accuracy. To conduct a more comprehensive analysis, we separately investigate the effects of ways and shots.

\begin{figure}[!t]
        \centering
        \includegraphics[scale=0.65]{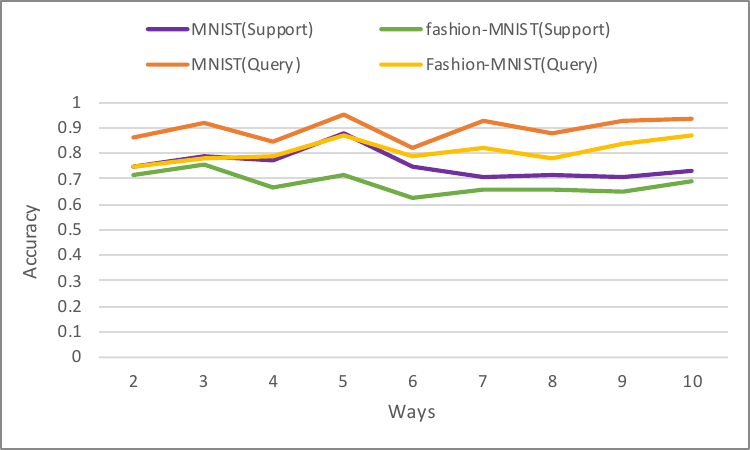}
        \caption{Investigating Attack Accuracy Across Data Ways: Surprisingly, the accuracy of the attack remains relatively stable despite variations in the number of data ways. This suggests that the attack performance is not significantly influenced by changes in the number of classes (ways), indicating a consistent level of accuracy across different class configurations.}
        \label{fig:Acc VS Ways}
\end{figure}

\begin{figure}[!t]
        \centering
        \includegraphics[scale=0.65]{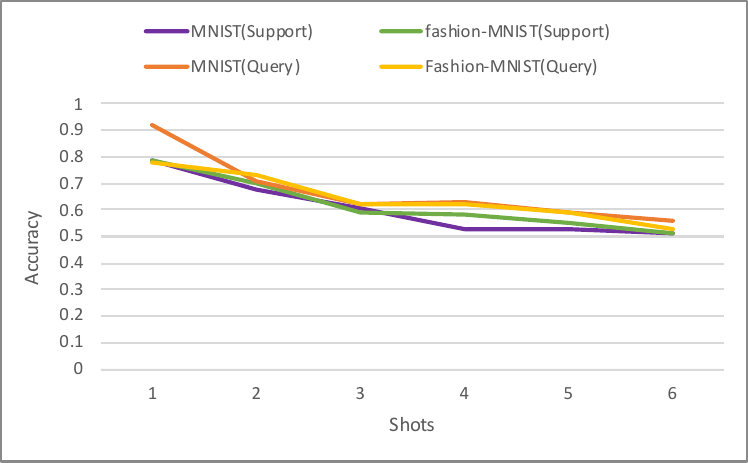}
        \caption{Investigating Attack Accuracy Across Data Shots: Notably, the accuracy demonstrates a decreasing trend as the number of data shots increases. This suggests that a higher number of data shots per class is associated with reduced accuracy in both support and query attacks.}
        \label{fig:Acc VS shots}
\end{figure}

In Figure \ref{fig:Acc VS Ways}, the trend of attack accuracy is depicted as the number of ways increases. The plot indicates that increasing the number of classes without a corresponding increase in samples from each class does not significantly impact attack accuracy. This is attributed to the fact that in the attack loss, each class has a separate contribution and is not affected by other classes, allowing the optimization to work separately for each class, resulting in no significant loss in accuracy with an increase in the number of classes. Additionally, in Figure \ref{fig:Acc VS shots}, we illustrate the effect of increasing shots on attack accuracy. As expected, with an increase in data samples from each class, the attack accuracy decreases. The reason behind this phenomenon lies in the fact that the gradient of loss is computed using an average over a batch of data. Therefore, increasing the data samples makes the reconstruction task more challenging. Although we still have a single gradient with a defined extent of information, we must now reconstruct more data from each class, contributing to the observed decrease in attack accuracy. This observation aligns with findings in previous research \cite{geiping2020inverting,zhu2019deep}.

\begin{figure}[!t]
        \centering
        \includegraphics[scale=0.65]{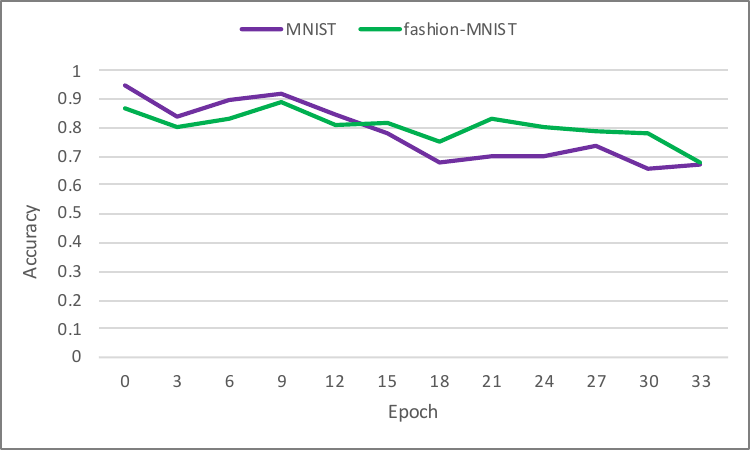}
        \caption{The Relationship Between Query Attack Accuracy and Training Epochs: Notably, the accuracy exhibits a decreasing trend as the model undergoes more training epochs. }
        \label{fig:training-query}
\end{figure}

The impact of model training on attack accuracy is another crucial factor to consider. The degree of model training varies significantly from the start to the end of the meta-learning process. Investigating the effect of the model training degree helps the meta-learner choose the optimal point to execute an attack with the highest probability of success. To investigate this, we perform the attack at different epochs of model training and analyze the results. In Figure \ref{fig:training-query}, it is evident that the query attack achieves the best results when the model is at epoch 0, indicating an untrained state. This aligns with prior research \cite{geiping2020inverting}, which suggests that the query attack, based on gradient reconstruction, is more effective when the model is untrained. However, it is essential to note that even after model training, the attack accuracy remains acceptable, indicating the susceptibility of the model to attacks even post-training.

\begin{figure}[!t]
        \centering
        \includegraphics[scale=0.65]{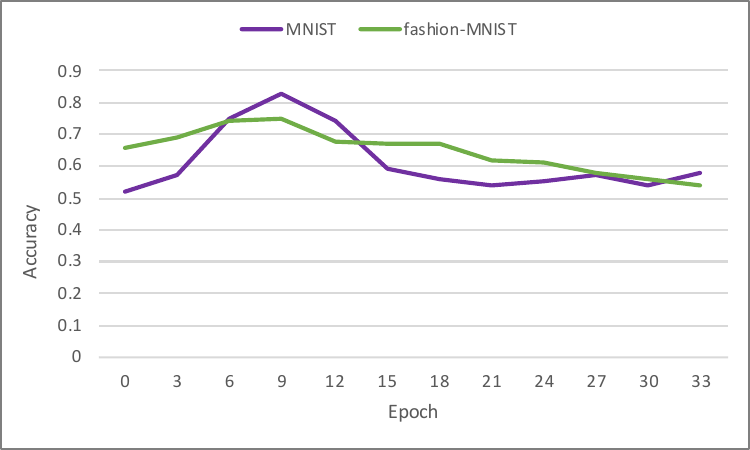}
        \caption{The Relationship Between Support Attack Accuracy and Training Epochs: Notably, the best attack accuracy is achieved when the model is partially trained.}
        \label{fig:training-support}
\end{figure}

Figure \ref{fig:training-support} illustrates the impact of training epochs on the support set attack. Interestingly, the pattern of accuracy variation with training epochs differs from that observed in the query attack. Here, the highest attack accuracy is achieved when the model is partially trained, indicating optimal results when the model has undergone some training but has not completed the training process. The rationale behind this observation lies in the behavior of loss gradients during training. Towards the end of training, loss gradients tend to become small and approach zero, containing insufficient information for effective model adaptation. Conversely, at the initial stages of training, the network learns general properties about classes and input images. At this point, the details of support set images have less impact on the model adaptation phase, making it challenging to differentiate between different input images. Consequently, the membership inference of specific data becomes more challenging, resulting in the best outcome when the model is partially trained.

\subsubsection{Efficiency}
The runtime of the attack is an important factor in assessing its performance in real-world applications. Two main factors influence attack runtime: the size of the data and the number of gradient descent iterations required for the attack. We evaluate the impact of these factors by conducting the attack under various settings, with the results presented in Table \ref{tab:runtime-support} for the support attack and Table \ref{tab:runtime-query} for the query attack. These experiments are performed on a simple personal PC equipped with a dual-core processor and 8 GB of RAM. Despite this modest setup, the attack runtimes are reasonable. As expected, an increase in data size and the number of optimization iterations leads to longer attack runtimes. It is also noteworthy that the query attack tends to be faster than the support attack, attributable to the fewer parameters that need to be optimized during the attack.

\begin{table}[]
\centering
\caption{Support Attack Runtimes(seconds)}
\label{tab:runtime-support}
\begin{tabular}{|c|ccccc|}
\hline
\multirow{2}*{Iteration} & \multicolumn{5}{c|}{Data Size}                                                                                            \\ \cline{2-6} 
                           & \multicolumn{1}{c|}{3}     & \multicolumn{1}{c|}{5}     & \multicolumn{1}{c|}{10}    & \multicolumn{1}{c|}{15}    & 20    \\ \hline
50                         & \multicolumn{1}{c|}{3.54}  & \multicolumn{1}{c|}{4.05}  & \multicolumn{1}{c|}{6.28}  & \multicolumn{1}{c|}{7.16}  & 8.84  \\ \hline
100                        & \multicolumn{1}{c|}{7.26}  & \multicolumn{1}{c|}{8.02}  & \multicolumn{1}{c|}{11.29} & \multicolumn{1}{c|}{14.57} & 17.45 \\ \hline
200                        & \multicolumn{1}{c|}{14.09} & \multicolumn{1}{c|}{15.93} & \multicolumn{1}{c|}{21.71} & \multicolumn{1}{c|}{29.71} & 34.79 \\ \hline
\end{tabular}
\end{table}

\begin{table}[]
\centering
\caption{Query Attack Runtimes(seconds)}
\label{tab:runtime-query}
\begin{tabular}{|c|lllll|}
\hline
\multirow{2}*{Iteration} & \multicolumn{5}{c|}{Data Size}                                                                                                          \\ \cline{2-6} 
                           & \multicolumn{1}{c|}{3}    & \multicolumn{1}{c|}{5}    & \multicolumn{1}{c|}{10}   & \multicolumn{1}{c|}{15}   & \multicolumn{1}{c|}{20} \\ \hline
50                         & \multicolumn{1}{l|}{1.34} & \multicolumn{1}{l|}{1.64} & \multicolumn{1}{l|}{1.94} & \multicolumn{1}{l|}{2.4}  & 3.42                    \\ \hline
100                        & \multicolumn{1}{l|}{2.56} & \multicolumn{1}{l|}{3.08} & \multicolumn{1}{l|}{3.86} & \multicolumn{1}{l|}{4.87} & 5.79                    \\ \hline
200                        & \multicolumn{1}{l|}{5.12} & \multicolumn{1}{l|}{6.04} & \multicolumn{1}{l|}{7.63} & \multicolumn{1}{l|}{9.22} & 14.11                   \\ \hline
\end{tabular}
\end{table}

\subsection{Validation of Privacy Preservation Mechanism}
Two key aspects need consideration to assess the effectiveness of adding noise to preserve model privacy. Firstly, it is essential to determine whether the attack accuracy decreases with the introduction of noise. Secondly, evaluating whether the meta-learning process remains effective in the presence of noise is crucial. Striking a balance between preventing privacy attacks and maintaining performance is crucial for the overall value of the privacy-preserving mechanism. 
For each noise addition method, we conduct an experiment by introducing different noise levels to the model and then monitoring the changes in model learning as well as the adversary's ability to attack the model. 

\begin{figure*}[!t]
    \centering

    \subfloat[]{
        \includegraphics[scale=0.35]{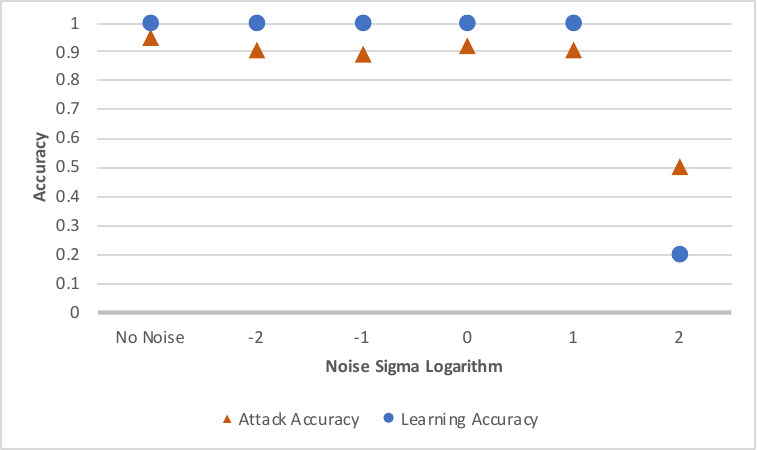}
        }
    \subfloat[]{
        \includegraphics[scale=0.35]{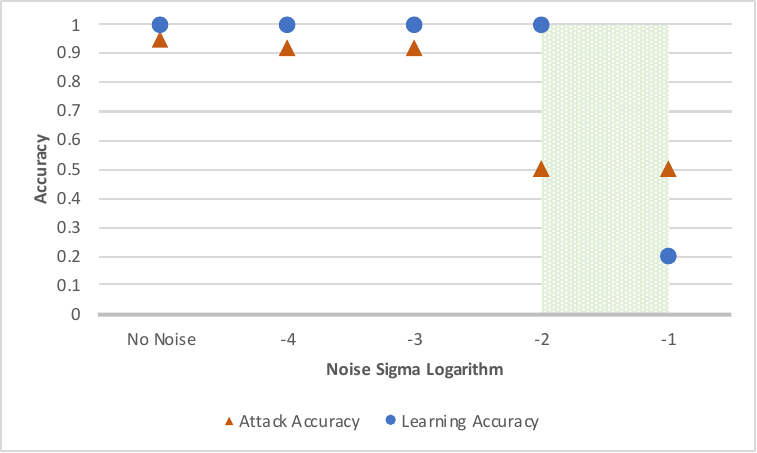}%
        }
    \subfloat[]{
        \includegraphics[scale=0.35]{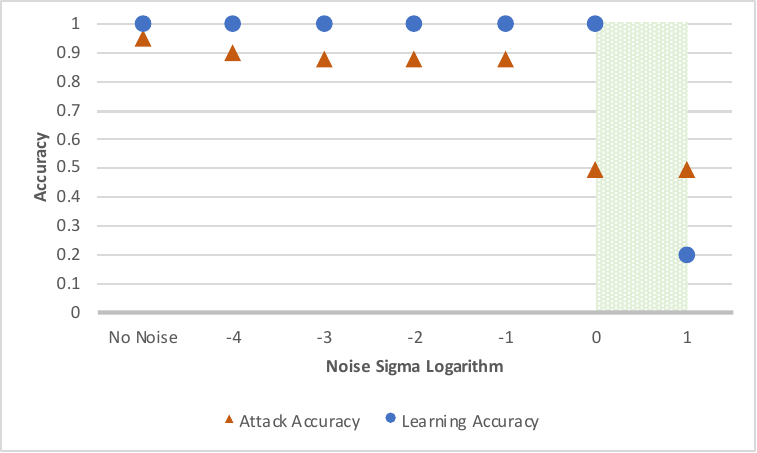}
        }
    \subfloat[]{
        \includegraphics[scale=0.35]{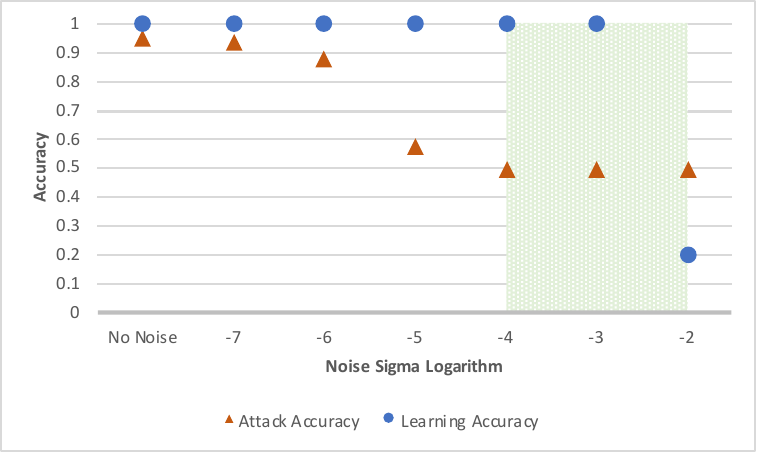}
        }
   
    \caption{Impact of Noise Addition on Query Attack (MNIST Dataset): Noise is added to (a) support data, (b) adaptation gradient, (c) query data, and (d) validation gradient, with varying effects observed on the attack performance and learning accuracy.}
    \label{fig:mnist-noise-query}
\end{figure*}

\begin{figure*}[!t]
    \centering
    \subfloat[]{
        \includegraphics[scale=0.35]{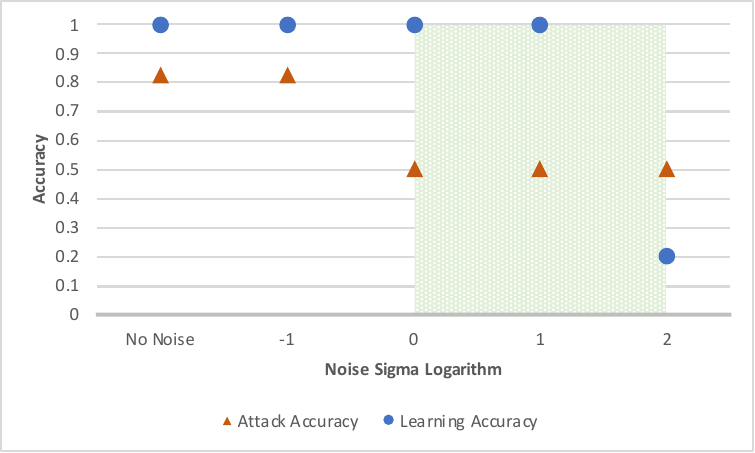}
        }  
    \subfloat[]{
        \includegraphics[scale=0.35]{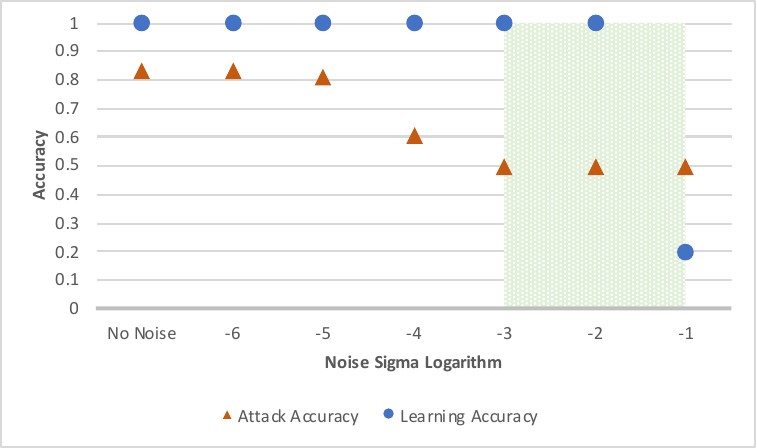}%
        }
    \subfloat[]{
        \includegraphics[scale=0.35]{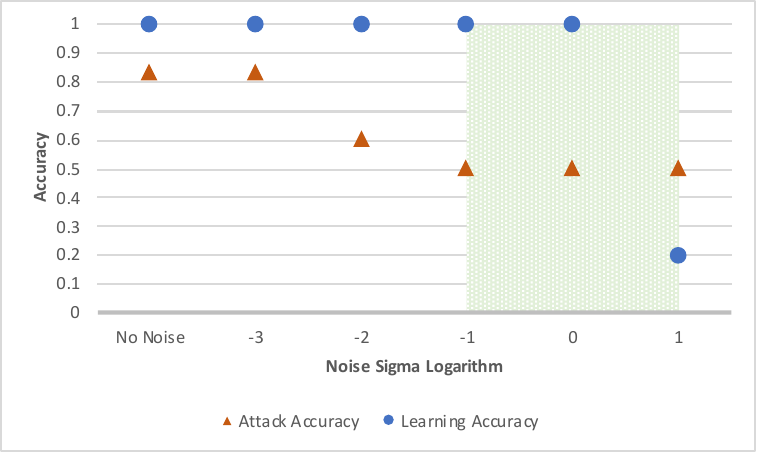}
        }        
    \subfloat[]{
        \includegraphics[scale=0.35]{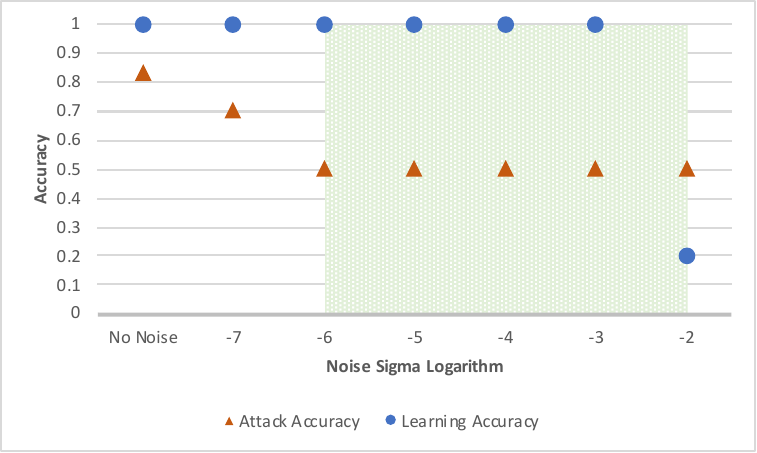}
        } 

    \caption{Impact of Noise Addition on Support Attack (MNIST Dataset): Noise is added to (a) support data, (b) adaptation gradient, (c) query data, and (d) validation gradient, with varying effects observed on the attack performance and learning accuracy. }
    \label{fig:mnist-noise-support}
\end{figure*}

\begin{figure*}[!t]
    \centering
     \subfloat[]{
        \includegraphics[scale=0.35]{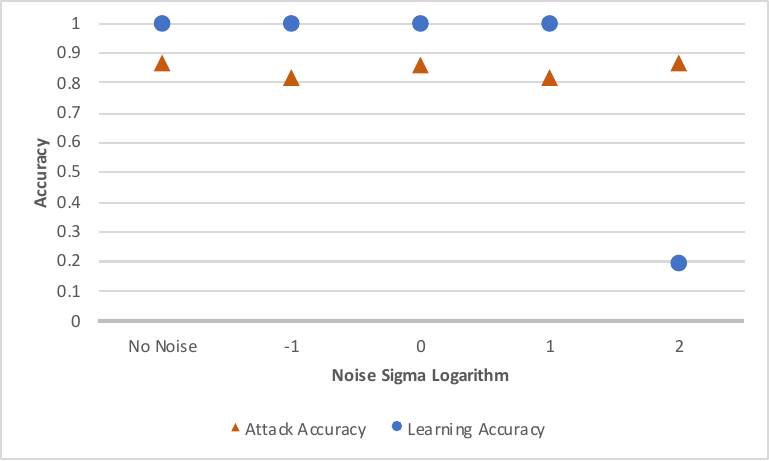}
        }        
    \subfloat[]{
        \includegraphics[scale=0.35]{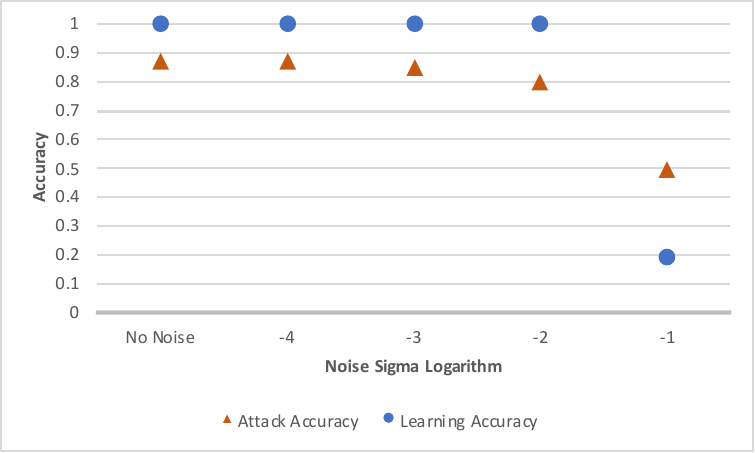}%
        } 
    \subfloat[]{
        \includegraphics[scale=0.35]{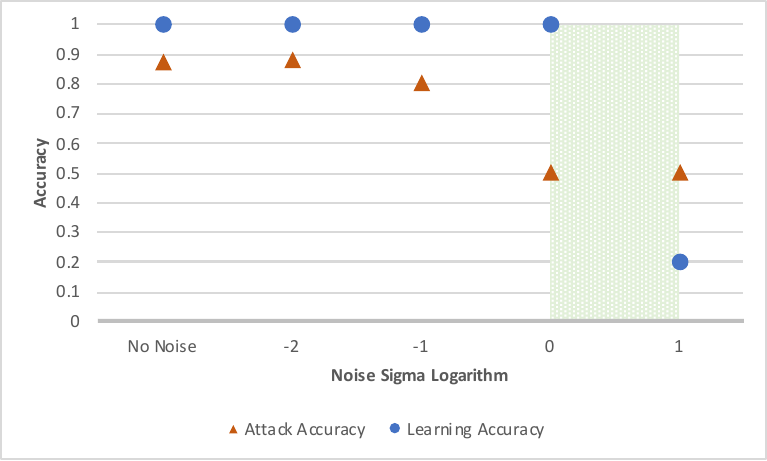}
        }       
    \subfloat[]{
        \includegraphics[scale=0.35]{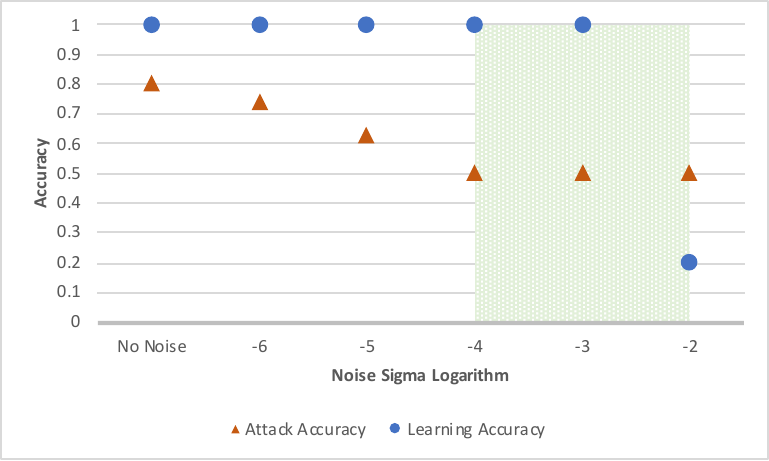}
        }

    \caption{Impact of Noise Addition on Query Attack (Fashion-MNIST Dataset): Noise is added to (a) support data, (b) adaptation gradient, (c) query data, and (d) validation gradient, with varying effects observed on the attack performance and learning accuracy. }
    \label{fig:fashion-noise-query}
\end{figure*}

\begin{figure*}[!t]
    \centering
    \subfloat[]{
        \includegraphics[scale=0.35]{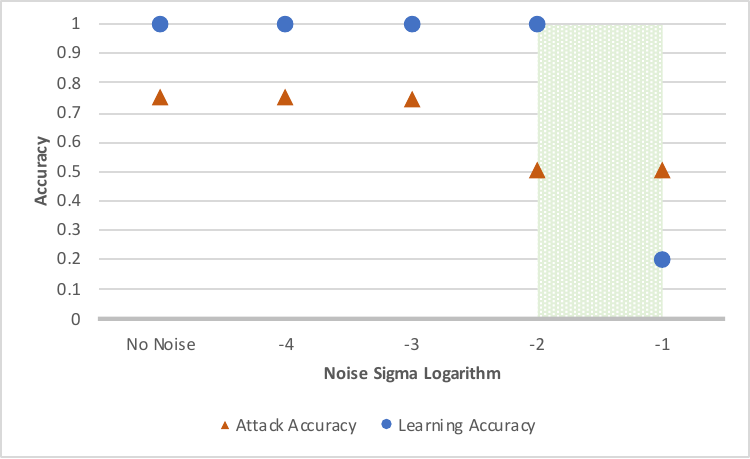}
        }      
    \subfloat[]{
        \includegraphics[scale=0.35]{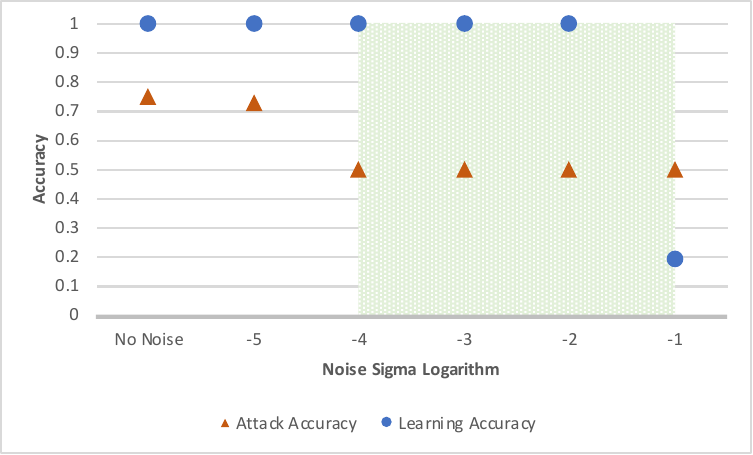}%
        }
     \subfloat[]{
        \includegraphics[scale=0.35]{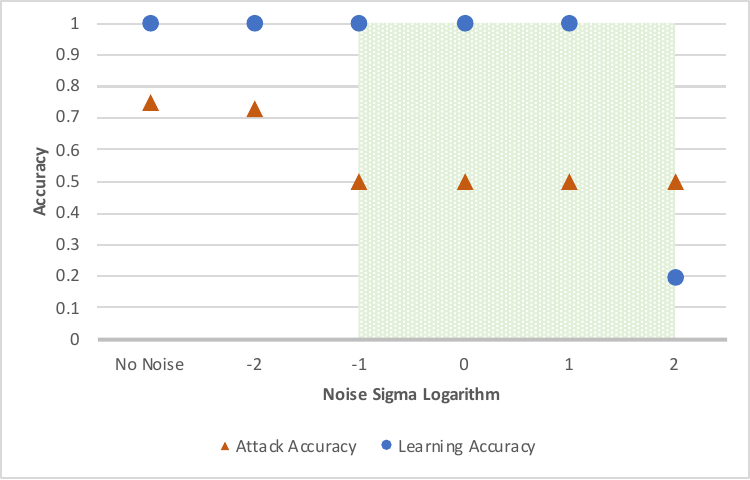}
        }    
    \subfloat[]{
        \includegraphics[scale=0.35]{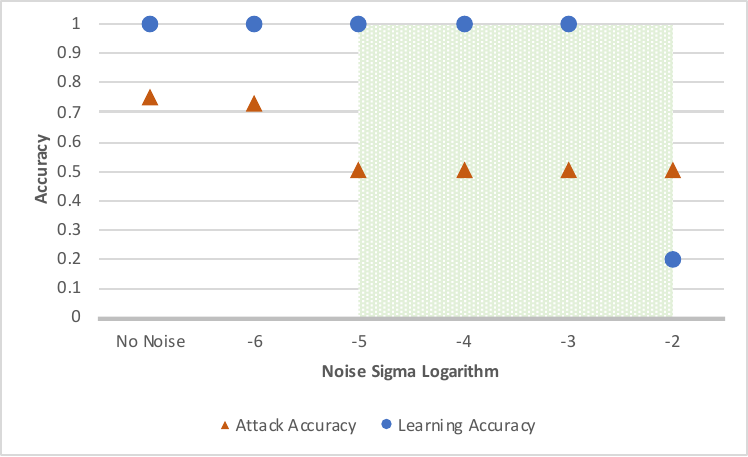}
        }
    
    \caption{Impact of Noise Addition on Support Attack (Fashion-MNIST Dataset): Noise is added to (a) support data, (b) adaptation gradient, (c) query data, and (d) validation gradient, with varying effects observed on the attack performance and learning accuracy. }
    \label{fig:fashion-noise-support}
\end{figure*}

In Figures \ref{fig:mnist-noise-query}, \ref{fig:mnist-noise-support}, \ref{fig:fashion-noise-query} and \ref{fig:fashion-noise-support} the blue points represent the model learning accuracy across various noise levels, while the red points indicate attack accuracy. The desired noise level is where the attack accuracy has decreased while the learning accuracy remains acceptable. This optimal region is highlighted in green. We refer to this area as the "confident noise area". Figure \ref{fig:mnist-noise-query} and \ref{fig:fashion-noise-query} displays the experiments for query attacks, and Figure \ref{fig:mnist-noise-support} and \ref{fig:fashion-noise-support} are for support attacks.

Interestingly, the confident noise areas are larger in support attacks, suggesting that this type of attack is more sensitive to noise addition, confirming the idea that support data is harder to be attacked. Additionally, it is notable that adding noise to the query data during the validation phase is more effective than adding noise to the support set. This observation further confirms the idea that the contribution of the adaptation phase on query data in the gradient shared with the meta-learner is more significant than the adaptation phase on the support set.

\section{Conclusion}
\label{sec:conclusoin}
In this study, we delved into data privacy within federated meta-learning, explicitly focusing on the MAML algorithm. Our analysis centered on the algorithm's shared gradient to assess the data leakage probability. We discovered that attacking the task data in MAML differs significantly and presents more challenges than conventional federated learning, necessitating the development of new attack methodologies. To address this, we proposed two distinct algorithms for conducting membership inference attacks on the MAML algorithm—one targeting the support set and the other the query set. Additionally, we introduced four different noise injection methods aimed at bolstering the algorithm's defenses against such attacks. Through our experiments, we demonstrated the efficacy of these attacks and subsequently explored the effectiveness of each noise injection method in safeguarding the algorithm against potential breaches.

\newpage
\bibliographystyle{IEEEtran}
\bibliography{refs}
\newpage
\appendix
\section{More Theoretical Analysis} \label{app1}
In Section \ref{sec:Attacks on MAML}, we elucidate Proposition 1, which reveals that, under certain conditions, the gradient shared by a task learner in the meta-learning algorithm lacks sufficient information about the support and query sets. The proposition establishes the existence of alternative support and query sets capable of generating the same gradient within this context. Here, we aim to demonstrate the existence of networks where these specified conditions hold.

Consider the following network, which consists of a single neuron. It takes an input $x \in \mathbf{R}$ and computes $z = ax + b$ as the neuron output, where $a, b \in \mathbf{R}$ denotes the weight and bias. The activation function is the Rectified Linear Unit (ReLU), expressed as $\hat{y} = \text{ReLU}(z)$. Given an input pair $(x, y)$ to the network, we compute the Mean Squared Error (MSE) loss, resulting in a loss value $J = \text{MSE}(y, \hat{y})$. The gradient $G$ and Hessian $H$ are then computed as follows:
\begin{equation}
G = \begin{bmatrix}
2x(\hat{y}-y)f_z(1) & 2(\hat{y}-y)f_z(1)
\end{bmatrix},
\end{equation}

where $f_y(x)$ is defined as:

\begin{equation}
f_y(x) = \begin{cases}
x & y \ge 0 \\
0 & y < 0
\end{cases},
\end{equation}

and the Hessian of the loss ($H$) is computed as:

\begin{equation}
H = \begin{bmatrix}
2x(x-y)f_{z}(1) & 2x(1-y)f_z(1)\\
2(x-y)f_{z}(1) & 2(1-y)f_z(1)
\end{bmatrix}
\end{equation}

Now, we can examine the two conditions outlined in Proposition 1. The invertibility of $I + \nabla_{\omega^2} \mathcal{L}(\omega,\omega,M^s)$, equivalent to $I+H$, can be verified by confirming that $|H+I|>0$, a condition easily met with suitable assignments of $x$ and $y$. Additionally, we can ascertain that $g (I + \nabla_{\omega^2} \mathcal{L}(\omega,\omega,M^s) )^{-1}$ contains at least one non-zero entry in each row. This follows from the fact that $g$ already possesses at least one non-zero entry in each row, making it sufficient to assign values to $x$ and $y$ such that $(I+H)^{-1}$ has all non-zero entries, which is achievable.
Therefore, the condition outlined in Proposition \ref{prop:prop1} can be satisfied in certain networks. This implies that in these networks, the MAML gradient $g$ is not unique for a specific $D^s$ and $D^q$. Consequently, we can choose $M^s$ and $M^q$ in such a way that they produce the identical gradient $g$.

\end{document}